\documentclass[final]{cvpr}

\usepackage{times}
\usepackage{epsfig}
\usepackage{graphicx}
\usepackage{amsmath}
\usepackage{amssymb}
 \usepackage{booktabs}
 \usepackage{multirow}
\usepackage{pifont}
\usepackage{caption}
\usepackage{subcaption}
\usepackage[pagebackref=true,breaklinks=true,colorlinks,bookmarks=false]{hyperref}

\def\cvprPaperID{****} 
\def\confYear{CVPR 2021}

\begin{document}

\title{Interpretable Deep Learning Classifier by Detection of Prototypical Parts \\ on Kidney Stones Images}

\author{Daniel Flores-Araiza$^{1}$, Francisco Lopez-Tiro$^{1}$, Elias Villalvazo-Avila$^{1}$, Jonathan El-Beze$^{2}$, \\ Jacques Hubert$^{2}$, Gilberto Ochoa-Ruiz$^{1}$, Christian Daul$^{3}$\\
\\
$^{1}$Tecnologico de Monterrey, School of Engineering and Sciences, Mexico\\
$^{2}$CHU Nancy, Service d’urologie de Brabois, Nancy, France\\
$^{3}$Centre de Recherche en Automatique de Nancy, Université de Lorraine, France \\
}

\maketitle

\begin{abstract}
Identifying the type of kidney stones can allow urologists to determine their formation cause, improving the early prescription of appropriate treatments to diminish future relapses. However, currently, the associated ex-vivo diagnosis (known as morpho-constitutional analysis, MCA) is time-consuming, expensive and requires a great deal of experience, as it requires a visual analysis component that is highly operator dependant. 
Recently, machine learning methods have been developed for in-vivo endoscopic stone recognition. Shallow methods have demonstrated to be reliable and interpretable but exhibit low accuracy, while deep learning-based methods yield high accuracy but are not explainable. 
However, high stake decisions require understandable computer-aided diagnosis (CAD) to suggest a course of action based on reasonable evidence, rather than to merely prescribe one. 
Herein, we investigate means for learning part-prototypes (PPs) that enable interpretable models.
Our proposal suggests a classification for a kidney stone patch image and provides explanations in a similar way as those used on the MCA method.

\end{abstract}


\begin{figure*}[ht] 
    \centering
    \subfloat[\small{Model architecture}]{
    \label{fig:a}
    \includegraphics[width=0.48\linewidth]{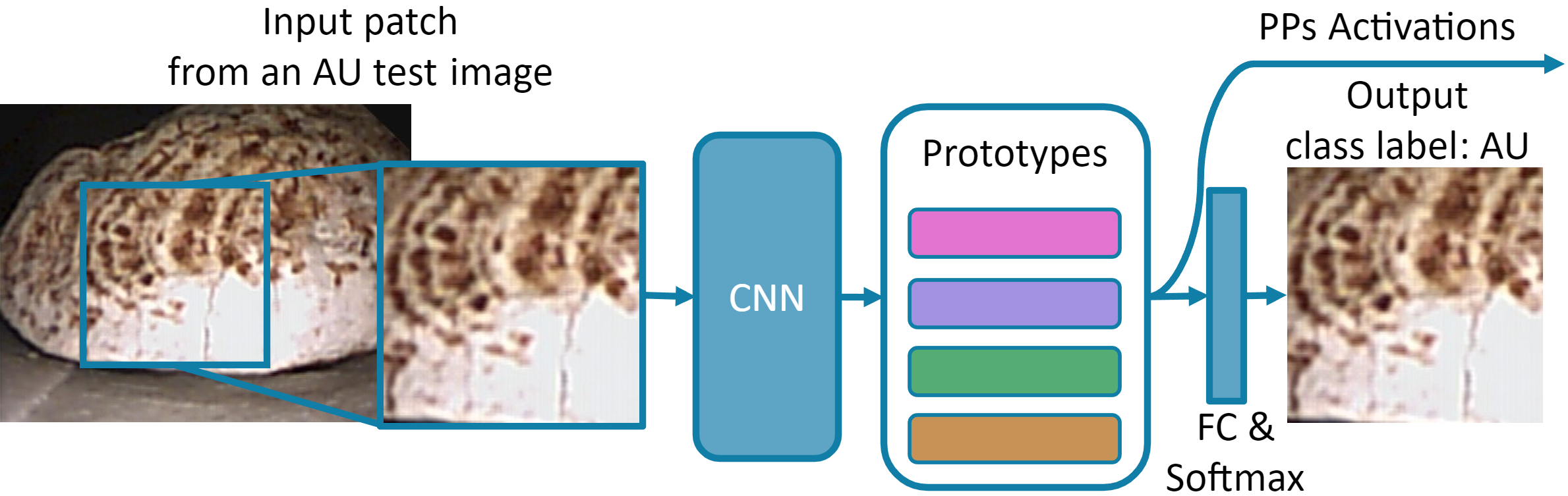}}
    \subfloat[\small{Explanations}]{
    \label{fig:b}
    \includegraphics[width=0.48\linewidth]{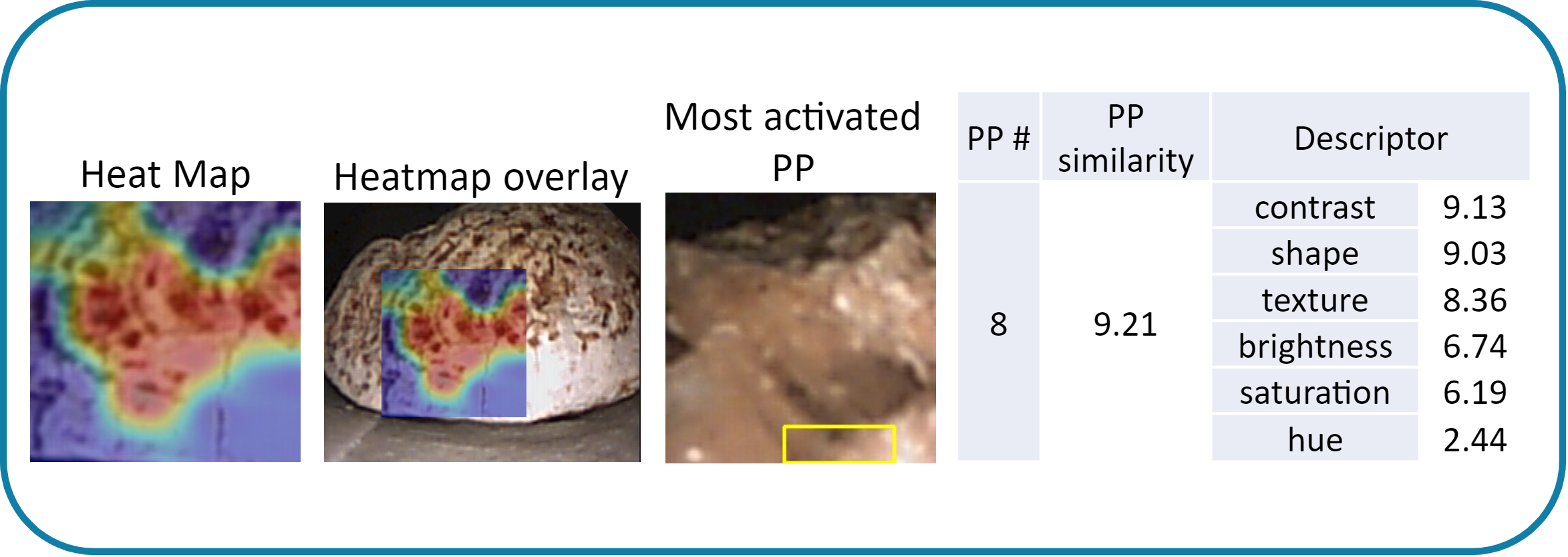}}
    \caption{\small{
    (a) Overall view of the proposal workflow, using ProtoPNet to obtain particular explanations for an input image. By use of PPs we provide explanations of the output classification, in tree different ways on (b), with a heatmap of the relevant parts on the input image, training images detected similar to the input, and measures of visual characteristics (descriptors) important for the activated PPs. 
    }}
\label{fig:MI_XAI_PPs_diagram}
    \end{figure*}
\section{Introduction}
\label{introduction}

Urolithiasis disease refers to the formation of kidney stones (KS). 
Several industrialized countries present a high incidence of kidney stone episodes (around 10$\%$ of the population is affected \cite{viljoen2019renal, nassir2018impact}). 
The stone formation is a multi-factorial process \cite{corrales2021classification, daudon2018recurrence}, where the diet is one of the most important factors   \cite{friedlander2015diet, siener2003fluid} but several complementary factors can produce it (e.g., hereditary-family history, chronic diseases, and sedentary lifestyle).
Early identification of the type  of kidney stone aids the urologist to have an accurate diagnosis, enabling them to prescribe the appropriate treatment (e.g., diet adaptation or surgery), but most importantly, to reduce eventual relapses \cite{friedlander2015diet}. 

The kidney stone type of an ex-vivo sample (extracted during endoscopic surgery) can be identified using a two-step procedure, a method known as Morpho-Constitutional Analysis (MCA) \cite{daudon2016comprehensive, daudon2018recurrence}. First, a microscopic morphological examination of the visual characteristics of the stone (e.g., size, form, color, texture, and appearance of the surface and section view) is strongly correlated with the molecular study \cite{corrales2021classification} and it enables to preserve important diagnostic information. On the other hand, an Infrared-spectrophotometry analysis leads to a more precise identification of the crystalline composition of the stone \cite{estepa1997contribution}. 

Although the MCA efficiently establishes the type of ex-vivo kidney stones, it is very difficult to provide a reliable diagnosis during an endoscopic intervention (the results of the MCA may take days). Also, it is time-consuming and tedious (fragments extraction can take up to one hour of surgery) \cite{corrales2021classification} and very difficult to train specialists on.
Recently, it has been suggested that an endoscopic (intra-operative) stone recognition (ESR)  CAD tool could help to obtain a faster diagnosis based solely on the video signal information provided by the endoscope, in tandem with the visual aid on the screen \cite{estrade2021towards}. Also, the operations are quicker to perform and less traumatic, due to dusting can fragment and destroy the kidney stones inside the urinary tract.

Several methods have been proposed in recent years for performing automated ESR, based both on traditional and deep learning techniques with very encouraging results \cite{ochoa2022vivo}. 
On the one hand, shallow Machine Learning (ML) models have shown that the efficient extraction of features (e.g., color and texture) in kidney stone views (surface and section) can have a significant impact on the classification (with accuracy $\ge90\%$) on in-vivo endoscopic images (strongly correlated to the morphological analysis of kidney stones) \cite{martinez2020towards, lopez2021assessing, ochoa2022vivo}. 
However, in visualizations such as UMAP \cite{mcinnes2018umap}, the clusters are not tight enough, which could mean that they are not the best features that could be used in the classifier.
On the other hand, Deep Learning (DL) based models \cite{lopez2021assessing, estrade2021towards, black2020deep, ochoa2022vivo} have shown excellent results (high accuracy $\ge95\%$) for extracting features relevant to the classifier (and tight clusters in UMAP). However, DL models lack interpretation of the features they extract, making these models not very useful in clinical settings. 

As a matter of fact, current ML and DL models are unable to describe the inner workings that led to a given prediction beyond the class label.
Therefore, these types of models cannot provide useful information to the specialist to understand how the input image was used to perform a diagnosis (i.e., classify the kidney stone type).

In order to pave the way for AI-based ESR using deep learning techniques, in this work, we leverage recent strides in explainability that seek to base image classifiers decisions on case-based reasoning to make them more interpretable \cite{ProtoPnet, PP_Descriptors, ProtoPshare, AIAIBL}. Additionally, we provide both visual explanations and quantitative information about visual characteristics deemed important by the network. It must be emphasized that our approach follows the reasoning processes of urologists in detecting morphological relevant features of each image (i.e., MCA). Overall, this work is aimed at facilitating human-machine collaboration in the context of CAD tools for urolithiasis prevention.

\section{Materials and Methods}
\label{materials_n_methods}

\subsection{Kidney stone dataset}
\begin{figure}[hbtp]
\centering
    \includegraphics[width=0.95\linewidth]{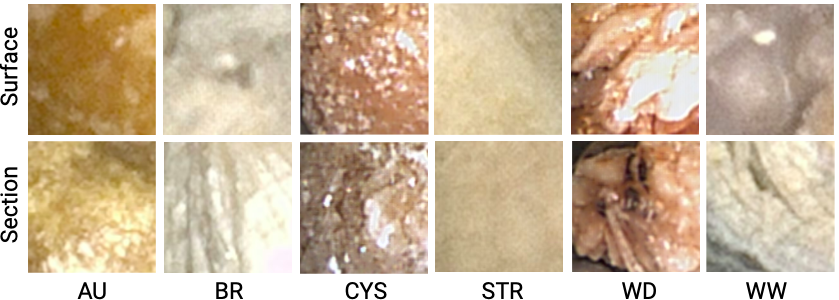}
    \caption{\small{Examples of ex-vivo kidney stones generated patches.}} 
\label{fig:im1}
\end{figure}

The ex-vivo dataset includes 305 kidney stone images acquired (two reusable digital flexible ureteroscopes from Karl Storz using video columns: Storz Image 1 Hub and Storz image1 S) and labeled manually by the urologist Jonathan El Beze$^{2}$ (for more details, see \cite{elbeze2022}).  
For this study, we make use of an ex-vivo image dataset divided in three subsets:  177 surface images, 128 section images and the third subset of 305 images (177 surface and 128 section images) of the six kidney stone types with the highest incidence: Acide Urique (AU), Brushite (BRU), and Cystine (CYS), Struvite (STR), Weddellite (WD), Whewellite (WW). Patches of this dataset are shown in Fig. \ref{fig:im1}. 

\begin{table*}[]
\centering
\caption{\small{Weighted average metrics comparison for section, surface, and mixed patches. ProtoPNet with VGG19 with batch normalization as the backbone (PPN-VGG19bn). VGG 19-layer model, configuration `E', with batch normalization (VGG19bn).
}}
\label{tab:my-table}
\scalebox{0.88}{\begin{tabular}{@{}ccccccccccccc@{}}
\toprule
\multirow{2}{*}{Model} & \multicolumn{3}{c}{Accuracy} & \multicolumn{3}{c}{Precision} & \multicolumn{3}{c}{Recall} & \multicolumn{3}{c}{F1 score} \\ \cmidrule(l){2-13} 
                       & Surface  & Section  & Mixed  & Surface   & Section  & Mixed  & Surface  & Section & Mixed & Surface  & Section  & Mixed  \\ \midrule
PPN-VGG19bn            & 0.98     & 0.99     & 0.97   & 0.98      & 0.99     & 0.97   &
 0.98     & 0.99    & 0.97  & 0.98     & 0.99     & 0.97   \\
VGG19bn                & 0.99     & 1.00     & 1.00   & 0.99      & 1.00     & 1.00   & 0.99     & 1.00    & 1.00  & 0.99     & 1.00     & 1.00   \\
AlexNet                & 0.96     & 0.97     & 0.97   & 0.96      & 0.97     & 0.97   & 0.96     & 0.97    & 0.97  & 0.96     & 0.97     & 0.97   \\ \bottomrule
\end{tabular}
}
\end{table*}

However, the identification of kidney stones is not usually performed on whole images \cite{serrat2017mystone, torrell2018metric, black2020deep, martinez2020towards}. Thus, patches of 200$\times$200 pixels (minimal size enable to capture enough texture and color information) were cropped from the original images to increase the size of the training dataset (for more details, see \cite{lopez2021assessing}). 
A total of 2000 patches are available per class (AU, BRU, CYS, STR, WD, and WW) and view (surface, section, and mixed). The train and test relationship were $80\%$ (38400 images) and $20\%$ (9600 images), respectively. 
In order to limit the over-fitting produced by the small size of the available training dataset, data augmentation was performed. Additional patches were obtained by applying geometrical transformations (patch flipping, and perspective distortions), increasing the number of training patches from 38400 to  1152000 using data augmentation. The patches were also ``whitened" using the mean $m_{i}$ and standard deviation $\sigma_{i}$ of the color values $I_{i}$ in each channel $(I^{w}_{i} = (I_{i}-m_{i}\sigma_{i})$, with $i=R,G,B)$. 

\subsection{ProtoPNet plus descriptors}\label{Methods}
By use of a Prototypical Part Network (ProtoPNet), able to identify several parts of an image, where it thinks a part of the image looks like a learned prototypical part of some class (as it can be seen in Fig. \ref{fig:a} and \ref{fig:b}). This type of model makes its prediction on a weighted combination of the similarity scores between parts of the image and the learned part-prototypes (PPs).
This capacity yields predictions that are relatively easy to understand by users, rendering it interpretable. 
We apply the methodology presented in previous works, that proposed and used ProtoPnet models \cite{ProtoPnet, ProtoPshare}.
However, PPs may still depend on non-apparent characteristics from the input image, 
reason for us to quantify the sensitivity of PPs to a set of perturbations \cite{PP_Descriptors}, which we call ``descriptors". These descriptors indicate why the classification model deemed an image patch and PP (part-prototype) similar.
It's worth noticing that PPs are vectors in latent space that should learn discriminative, prototypical parts of a class. Thus, high dimensional reduction projections, UMAP visualizations as an example, contain global information of the main characteristics of the whole ProtoPNet model. 

\subsection{Implementation details} \label{Implementation_Details}

The ProtoPNet architecture consists of a standard Convolutional Neural Network (CNN, e.g. ResNet),
followed by a prototype layer and a fully-connected layer. 
The prototype layer consists of a pre-determined number of class-specific prototypes. Herein, we use 10 part-prototypes per class, the initialization and training procedure for our training also follows \cite{ProtoPnet, PP_Descriptors}, using a pre-trained VGG19 (with batch normalization) on ImageNet as CNN backbone.

\begin{figure}[ht] 
    \centering
    \includegraphics[width=0.9\linewidth]{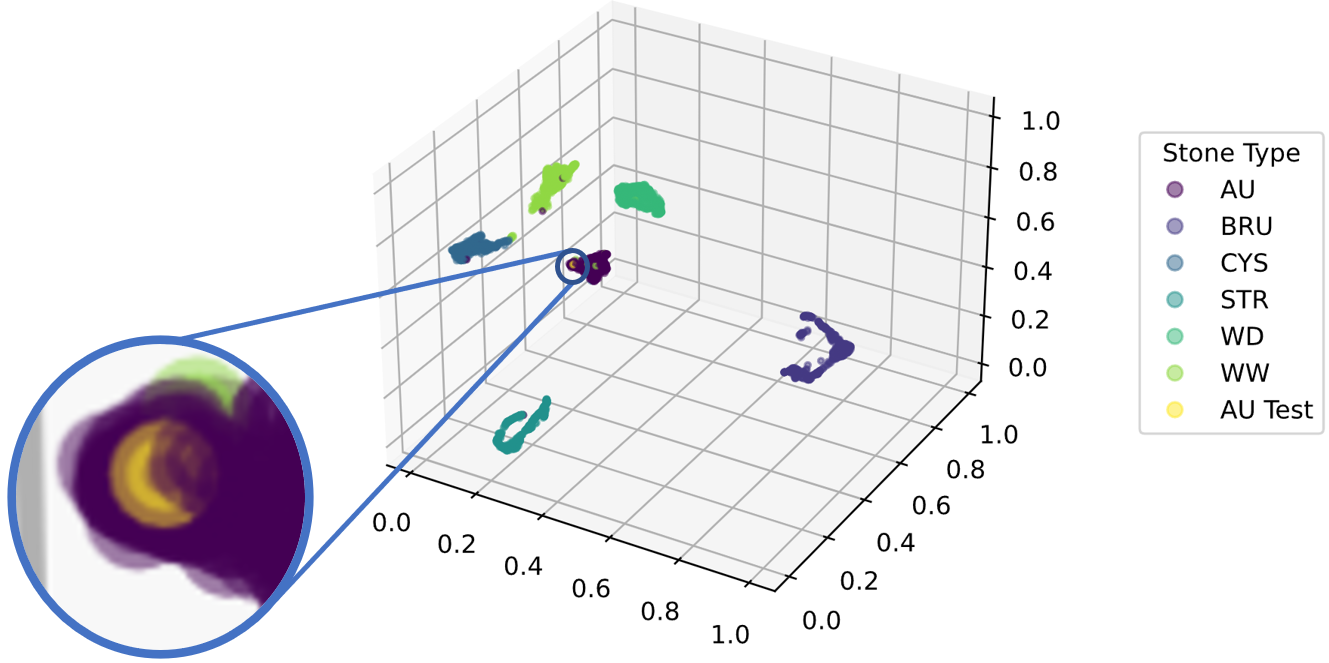}
    \caption{\small{
    UMAP of the Part-Prototypes activations on section images. Our approach allows obtaining separate clusters of the output classes. 
    On the zoomed example (blue circle) can be appreciated the projection of a new classification (yellow point) is surrounded by samples of the same class (purple points), indicating a high certainty on the correct classification of the test sample.
    }}
    \label{fig:umaps}
\end{figure}
    
\section{Results and Discussion}\label{Results}

Evaluation metrics of our model, its backbone model, and AlexNet (as reference), are reported in Table \ref{tab:my-table}. The performance of ProtoPNet is comparable with its corresponding uninterpretable backbone model ($\le3\%$ difference).
In contrast to black-box classifiers, our proposal provides explanations for input images, by showing the activation area of PPs, the corresponding representative image to each activated PPs, and their descriptors (Fig. \ref{fig:b}).

We plot PPs and their descriptors activations for input images on a UMAP visualization of the three most discriminant dimensions (umap1 to umap3). This UMAP allows seeing class separability for each output class of the ProtoPNet, as shown in Fig. \ref{fig:umaps}. 
In this way, additional global insight is gained in the case of a new classification observed surrounded by samples of the same class, which provides confidence of a correct classification (also described in Fig. \ref{fig:umaps}).
However, it was observed that multiple PPs end up indicating the same training patch as their explanation, a behavior similar to mode collapse on Generative Adversarial Networks (GANs), which limits the variety of possible explanations provided for an output.
The use of descriptors mitigates the cases for visually similar PPs by providing details on the characteristics most relevant for each PP \cite{PP_Descriptors}. 

\section{Conclusion and Future work}\label{Conclusions}

We showed that by training of PPs and extracting their descriptors we convert an uninterpretable VGG19 into an interpretable model.
This can facilitate the use of these models for ESR by a urologist. 
However, mode collapse of the learned PPs is a limitation on the current implementations of ProtoPNets. To prevent this, better initialization procedures and loss function adjustments will be explored.
Finally, we found indications of better class separability by use of PPs and their descriptors, to the point UMAP visualizations could be used to provide global context of the certainty of the output classification of a particular image. 

\section*{Acknowledgments}
The authors wish to thank the AI Hub and the CIIOT at ITESM for their support for carrying the experiments reported in this paper in their NVIDIA's DGX computer.

{\small
\bibliographystyle{ieee_fullname}
\bibliography{references}
}
\end{document}